\documentclass[letterpaper, conference]{IEEEtran}
\IEEEoverridecommandlockouts

\usepackage{cite}
\usepackage{amsmath,amssymb,amsfonts}
\usepackage{algorithmic}
\usepackage{graphicx}
\usepackage{textcomp}
\usepackage{xcolor}
\usepackage{listings}
\usepackage{float} 
\usepackage{pythonhighlight}
\usepackage[T1]{fontenc}
\usepackage{verbatim}
\usepackage{url}
\usepackage{amsmath}
\usepackage{multirow}
\usepackage{booktabs} 
\usepackage{wrapfig}
\usepackage{flushend} 
\usepackage{subcaption} % 或 subfigure
\newcommand{\etal}{\textit{et al}. }
\def\BibTeX{{\rm B\kern-.05em{\sc i\kern-.025em b}\kern-.08em
    T\kern-.1667em\lower.7ex\hbox{E}\kern-.125emX}}

\begin{document}

\title{Human-centric Reward Optimization for Reinforcement Learning-based Automated Driving using Large Language Models}

\author{
\IEEEauthorblockN{
Ziqi Zhou\IEEEauthorrefmark{1},
Jingyue Zhang\IEEEauthorrefmark{1},
Jingyuan Zhang\IEEEauthorrefmark{1},
Yangfan He\IEEEauthorrefmark{2},\\
Boyue Wang\IEEEauthorrefmark{3},
Tianyu Shi\IEEEauthorrefmark{4},
Alaa Khamis\IEEEauthorrefmark{5}\IEEEauthorrefmark{6}
}

\IEEEauthorblockA{\IEEEauthorrefmark{1}Faculty of Applied Science \& Engineering, University of Toronto, Toronto, Canada\\
Emails: \{zq.zhou, jingyuezjy.zhang, jymike.zhang\}@mail.utoronto.ca}

\IEEEauthorblockA{\IEEEauthorrefmark{2}Department of Computer Science, University of Minnesota -- Twin Cities, MN, USA\\
Email: he000577@umn.edu}

\IEEEauthorblockA{\IEEEauthorrefmark{3}Department of Mechanical Engineering, University of Wisconsin -- Madison, WI, USA\\
Email: bwang367@wisc.edu}

\IEEEauthorblockA{\IEEEauthorrefmark{4}Transportation Research Institute, University of Toronto, Toronto, Canada\\
Email: ty.shi@mail.utoronto.ca}

\IEEEauthorblockA{\IEEEauthorrefmark{5}IRC for Smart Mobility \& Logistics, 
King Fahd University of Petroleum and Minerals, Dhahran, Saudi Arabia\\
Email: alaa.rashwan@kfupm.edu.sa}

\IEEEauthorblockA{\IEEEauthorrefmark{6}Senior Member, IEEE}
}

\maketitle

\begin{abstract}
One of the key challenges in current Reinforcement Learning (RL)-based Automated Driving (AD) agents is achieving flexible, precise, and human-like behavior cost-effectively. This paper introduces an innovative approach that uses large language models (LLMs) to intuitively and effectively optimize RL reward functions in a human-centric way. We developed a framework where instructions and dynamic environment descriptions are input into the LLM. The LLM then utilizes this information to assist in generating rewards, thereby steering the behavior of RL agents towards patterns that more closely resemble human driving. The experimental results demonstrate that this approach not only makes RL agents more anthropomorphic but also achieves better performance. Additionally, various strategies for reward-proxy and reward-shaping are investigated, revealing the significant impact of prompt design on shaping an AD vehicle's behavior. These findings offer a promising direction for the development of more advanced, human-like automated driving systems. Our experimental data and source code can be found here\footnote{GitHub:~\url{https://github.com/JingYue2000/In-context_Learning_for_Automated_Driving}.}.

\end{abstract}


\section{Introduction}
\label{sec:intro}

With the advancement of artificial intelligence, an increasing number of machine learning techniques are being applied in the field of assisted and automated driving~\cite{AutonomousDriving},~\cite{Machine-learning-based-automatic-control}. Among them, Reinforcement Learning (RL) is a branch of machine learning that enables intelligent agents to learn optimal or near-optimal behaviors by interacting with their environments and receiving feedback in the form of rewards or penalties~\cite{Khamis2023}. RL showcases robust learning and adaptability, particularly in real-time decision-making scenarios prevalent in automated driving~\cite{1_Reinforcement_learning_An_introduction,Deep_Reinforcement_Learning_for_Autonomous_Driving:A_Survey,Autonomous_driving_planning}. 

Despite its potential benefits in enhancing safety~\cite{denseSafety}, efficient navigation, and reducing traffic congestion in real-world driving~\cite{comfortable_driving}, there are challenges hindering its widespread adoption~\cite{Survey_LLM_AD}. For example, one notable challenge is designing accurate reward functions that encourage safe, efficient, and human-like driving behavior; yet, there are other difficult aspects to consider as well:

\begin{itemize}
\item Perfecting RL agents to mirror human driving while ensuring safety and efficiency remains a complex task.
% \item High-speed environments raise safety worries.
\item Creating fair and easy-to-understand reward functions remains challenging due to driving task and environment complexity. 
\item  Traditional methods, heavily reliant on extensive labeled data for designing reward functions, often fail to adapt to variable user needs or evolving objectives in new contexts.
\end{itemize}

To solve these challenges, a shift in approach is emerging with In-context Learning (ICL),~\cite{ICT}. In recent years, In-Context Learning (ICL) has gained substantial interest and attention within the research community. ICL represents the ability of large language models to perform tasks by interpreting examples or instructions provided in the input prompt, without requiring updates to the model’s parameters. Unlike traditional training methods, ICL does not rely on large amounts of training data or parameter updates. Instead, it uses a small amount of contextual information and examples to help the model make decisions. In the context of automated driving, ICL offers a transformative solution by enabling LLMs to interpret natural language cues and generate reward signals that guide the behavior of reinforcement learning agents. This approach hypothesizes that having been trained by a large amount of content with a wide range of topics, including driving knowledge, LLMs were endowed with the ability to model a diverse set of learned concepts and the user could use the prompt to “locate” a previously learned concept~\cite{ai_stanford_blog}. By specifying an objective with a natural language prompt, LLM can convert user goals into reward signals and intuitively guide RL agents to desired outcomes \cite{5_Reward_design_with_language_models}. This innovative interaction between LLMs and RL agents not only advances opportunities in fields like automated driving, but also highlights the versatility of LLMs in complex decision making.

Building on the previously described theory, we present our approach to leveraging the synergy between LLMs and RL agents in automated driving scenarios. Specifically, we examine how LLM-guided reward signals influence RL training and performance, enabling the development of diverse and adaptive driving behaviors. Our proposed framework combines LLMs and RL to achieve more flexible and human-like decision-making in automated driving. By incorporating human-optimized reward designs, the RL agent learns driving behaviors that closely mimic human tendencies. This approach addresses complex decision-making challenges in dynamic, real-world scenarios, enhancing the adaptability and responsiveness of self-driving systems. The integration of LLMs and RL thus offers a promising pathway for advancing automated driving technologies. The contributions of this paper are summarized as follows:

\begin{itemize}
\item We proposed an RL training framework incorporating LLMs to simplify the RL-AD agents' reward design in AD scenarios, quoted as LLM-RL agent in this paper. % naming
\item We explore the integration of LLMs with RL agents to enable different human-like driving styles, allowing for dynamic adaptation.
\item We conducted a comparative analysis of previous approaches and showed that the integration of LLMs into the RL training process leads to more objective-aligned learning outcomes. 
\end{itemize}

The remainder of this paper is organized as follows. An overview of the related work is provided in Sections \ref{sec:review}. Section \ref{sec:problem} presents the problem formulation followed by describing the proposed approach in Section\ref{sec:approach}. The experiments and results are discussed in Sections \ref{sec:experiment} and \ref{sec:evaluation} and ref{sec:evaluation}. Finally, the conclusion and future work are summarized in Section \ref{sec:conclusion}.

\section{Related Work}
\label{sec:review}
\subsection{Large Language Models in Reinforcement Learning}
Large Language Models (LLMs), known for their capacity to learn from minimal context and deliver human-like responses following training with human data~\cite{LLMsurvey}, are increasingly considered as a potential solution when combined with Reinforcement Learning (RL) agents to perform higher-level tasks. In some studies, RL is employed to fine-tune existing LLMs\cite{llm_rl_1a, llm_rl_1b, llm_rl_1c}. Conversely, other research integrates RL with LLMs to execute specific tasks or develop particular skills\cite{llm_rl_3a, llm_rl_3b, llm_rl_3c}. In contrast, our research diverges from these methods by using LLMs as proxy rewards, enhancing RL model training as described in\cite{llm_rl_review}. 

The research landscape in using LLMs as proxies for RL rewards encompasses varied approaches. Kwon \etal pioneered the use of LLMs for learning in contexts such as the Ultimatum Game and negotiation tasks, focusing solely on LLM-generated rewards. \cite{5_Reward_design_with_language_models}. In contrast, our study employs a diverse set of rewards in dynamic automated driving system scenarios, broadening the scope of possibilities. Other researchers, like Xie \etal and Ma \etal \cite{xie2023text2reward},~\cite{ma2023eureka}, have explored reward design through direct Python code generation. Meanwhile, Song \etal in \cite{song2023selfrefined} required human inputs to iteratively fine-tune rewards, whereas another study \cite{Adaptive_Reinforcement_Learning_with_LLM-augmented_Reward_Functions} used LLM outputs as auxiliary rewards. Unlike these approaches, our method directly employs LLM responses to generate reward signals, simplifying the process and eliminating the need for human intervention.

\subsection{Large Language Models in Automated Driving}
Utilizing Large Language Models in automated driving enhances decision-making by leveraging their vast knowledge and reasoning capabilities. Research in vehical planning and control  primarily focuses on two approaches: prompt engineering~\cite{Dilu,Drive_Like_A_Human} and fine-tuning pre-trained models~\cite{LLM4D,Driving_with_LLMs}.

To address closed-loop driving tasks, Fu \etal~\cite{Drive_Like_A_Human} utilize LLMs to comprehend and interact with environments. Wen~\etal~\cite{Dilu}  extended this framework by incorporating reflection modules to aid LLMs in decision-making.
Our approach simplifies this process by considering only the ego car's most recent action. Similarly, 
Cui \etal ~\cite{Drive_As_Speak} proposed a method that provides the LLM with comprehensive environmental information collected from various vehicle modules, ensuring safer results from LLM decisions. DHowever, these methods often suffer from real-time latency issues.

Other researchers have focused on fine-tuning the pre-trained model. For example, Chen \etal~\cite{Driving_with_LLMs} fine-tuned LLMs using 10,000 generated driving scenarios, while DriveMLM~\cite{DriveMLM} combined linguistic processing with vehicle control to develop a more advanced system. In contrast, our work leverages LLMs to enhance RL performance, meeting real-time requirements by dynamically querying LLMs based on the current driving scenario.

\section{Problem Formulation and Modeling}
\label{sec:problem}

\subsection{Markov Decision Process Formalization}
An automated driving scenario can be modeled as a Markov Decision Process: $M = \langle S, A, \rho, R, \gamma \rangle$, where $S$ is the state space representing various driving scenarios (e.g., traffic density, relative speeds of vehicles, lane positions, etc.). $A$ is the action space representing the set of all possible driving actions (e.g., accelerate, decelerate, change lanes, etc.). $\rho : S \times A \times S \rightarrow [0, 1]$ is the transition probability function representing the likelihood of transitioning from one state to another given an action. $R : S \times A \rightarrow \mathbb{R}$ is the traditional reward function mapping states and actions to real-valued rewards. $\gamma$ is the discount factor, indicating the agent’s consideration for future rewards. The RL process can be defined within a Markov Decision Process (MDP) framework:

\begin{equation}
\pi^* = \arg\max_\pi \mathbb{E} \left[ \sum_{t=0}^{\infty} \gamma^t R(s_t, a_t) \mid \pi \right]
\end{equation}

where \( \pi \) is the policy can be seen as the agent’s brain or the decision-making strategy of the agent or the mapping from states or observations to actions.

Inspired by \cite{5_Reward_design_with_language_models}, we employed an LLM as the proxy for reward calculations in our framework to achieve a more interpretable and adaptable reward mechanism.
\begin{figure}[ht]
\centerline{\includegraphics[width=1\linewidth]{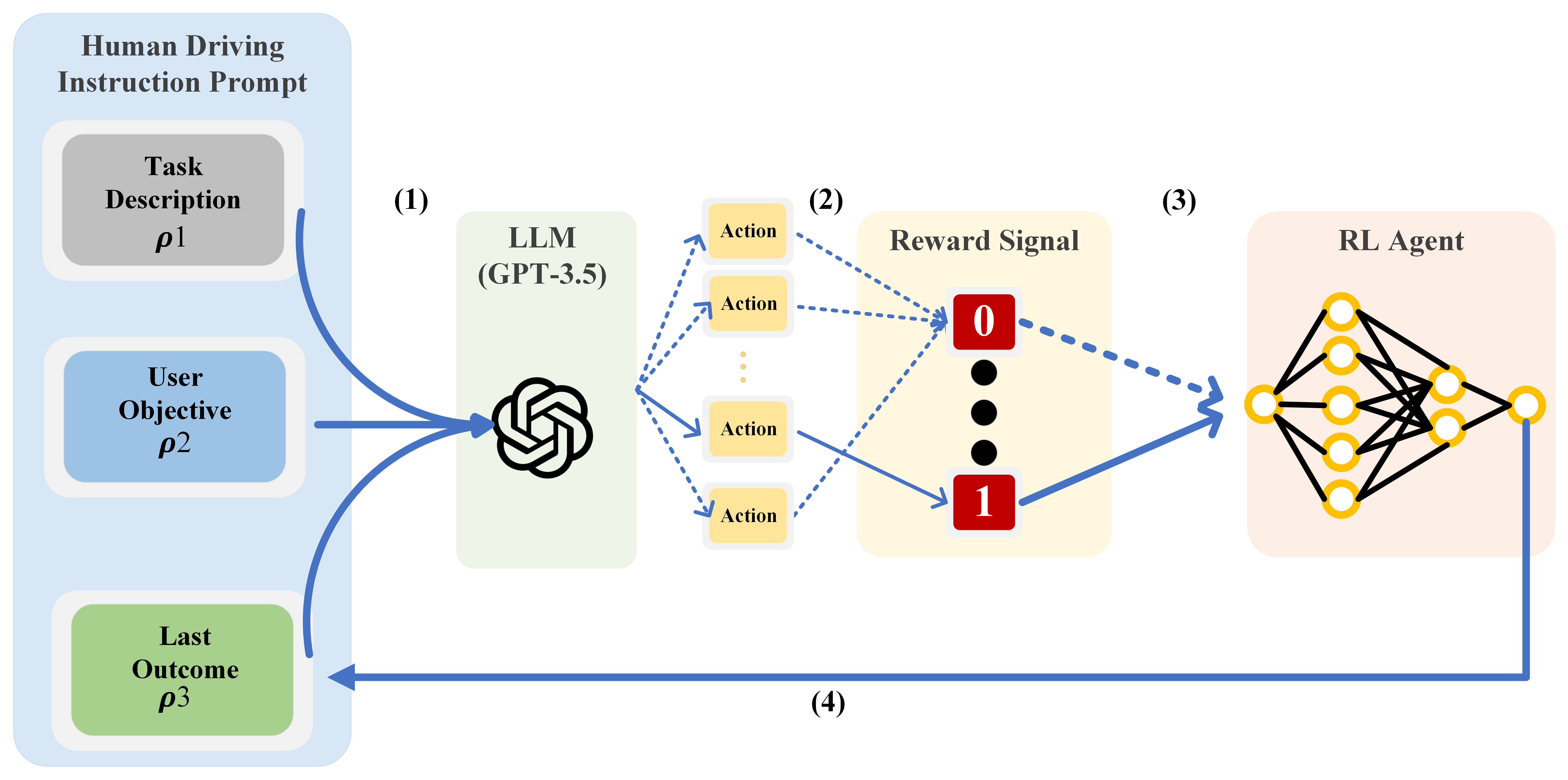}}
\caption{LLM acts as a reward proxy.}
\label{Framework When LLM Acts As the Reward Proxy}
\end{figure}
As illustrated in Fig.~\ref{Framework When LLM Acts As the Reward Proxy} our integrated framework uses the LLM to interpret textual prompts $\rho$ as input and outputs a reward signal. The text prompt $\rho$ consists of three major components: Task Description $\rho_1$, User Objective $\rho_2$ and Last Outcome $\rho_3$. The LLM is modeled as a function LLM: $A* \rightarrow A*$, taking the concatenated prompt $\varrho$ as input and outputting a string. A parser $A* \rightarrow {0, 1}$ is then used to map the textual output of the LLM to a binary reward signal, which refers to the (2) of Fig.~\ref{Framework When LLM Acts As the Reward Proxy}. Then, the binary reward signal will be sent to the RL agent for further training, which refers to the (3) of Fig.~\ref{Framework When LLM Acts As the Reward Proxy}. This framework replaces the traditional reward function $R$ with a proxy reward function LLM and can be used with any RL training algorithm. In this framework, LLMs are used to create a proxy reward function:

\begin{equation}
R_{\text{LLM}}(s, a) = f_{\text{LLM}}(\text{prompt}(s, a))
\end{equation}

To explore whether LLM-assisted agents could achieve a more balanced and safe driving behavior, emulating human-like decision-making in these scenarios, we implemented our framework in highway scenario as explained in Section \ref{sec:approach}.

\subsection{Reward Function Shaping Strategies}
To guide the agent's behavior, we introduce a combined reward function that integrates various factors contributing to safety, efficiency, and LLM-based evaluation.  The total reward function is expressed as:

\begin{equation}
R_{\text{total}}(s, a) = \alpha R_{\text{safety}}(s, a) + \beta R_{\text{efficiency}}(s, a) + \gamma R_{\text{LLM}}(s, a)
\end{equation}

where \( \alpha, \beta, \) and \( \gamma \) are weights that balance the contributions of each reward component, and their sum is normalized to 1. $\alpha$ + $\beta$ + $\gamma$ = 1, \quad \text{where } $\alpha$, $\beta$, $\gamma \geq 0$. Each category is defined in detail as follows:
\paragraph{Safety Reward (\(R_{\text{safety}}\))}
The safety reward focuses on factors that promote safe driving behavior, such as lane discipline and collision avoidance. It is defined as:
\begin{equation}
\begin{split}
R_{\text{safety}}(s, a) = &\ \lambda_1 \cdot \text{Lane\_discipline}(s, a) \\
&+ \lambda_2 \cdot \text{Collision\_avoidance}(s, a)
\end{split}
\end{equation}
where \( \text{Lane\_discipline}(s, a) \) rewards the agent for maintaining proper lane position, and \( \text{Collision\_avoidance}(s, a) \) penalizes the agent for actions leading to potential collisions. \( \lambda_1, \lambda_2 \) are the coefficients that adjust the impact of each safety-related factor.

\paragraph{Efficiency Reward (\(R_{\text{efficiency}}\))}
The efficiency reward aims to optimize fuel consumption and maintain appropriate speed. It is defined as:

\begin{equation}
\begin{split}
R_{\text{efficiency}}(s, a) = &\ \mu_1 \cdot \text{Fuel\_efficiency}(s, a) \\
&+ \mu_2 \cdot \text{Speed\_maintenance}(s, a)
\end{split}
\end{equation}
where \( \text{Fuel\_efficiency}(s, a) \) rewards the agent for minimizing fuel usage, and \( \text{Speed\_maintenance}(s, a) \) encourages the agent to maintain a speed that is efficient and within safe limits. \( \mu_1, \mu_2 \) are the coefficients for each efficiency factor.

\paragraph{LLM Reward (\(R_{\text{LLM}}\))}
The LLM reward leverages the evaluation provided by a large language model, which assesses the appropriateness of the agent's actions based on human-like driving patterns. This reward is binary (either 0 or 1), defined as:
\begin{equation}
R_{\text{LLM}}(s, a) = \nu \cdot \text{LLM\_feedback}(s, a)
\end{equation}
where \( \text{LLM\_feedback}(s, a) \) provides a 0 or 1 reward depending on whether the LLM deems the action \( a \) in state \( s \) to be appropriate, and \( \nu \) is the weight for this reward.
\paragraph{Adjusted Total Reward}
We use a tangent function to scale the combined reward to the [0,1] range:
\begin{equation}
\begin{split}
R_{\text{ad}}(s, a) = \sigma \Big( 
    &\alpha R_{\text{safety}}(s, a) \\
    &+ \beta R_{\text{efficiency}}(s, a) \\
    &+ \gamma R_{\text{LLM}}(s, a) 
\Big)
\end{split}
\label{eq : total_reward}
\end{equation}
where \( \sigma \) is the scaling function ensuring the reward remains within the [0,1] range. This adjusted reward is then used to guide the agent's decision-making process.
\paragraph{Environment Reward}
In addition to the main reward components, we include an environment-based reward that encourages speed optimization while penalizing collisions:

\begin{equation}
\text{Env\_reward}(a) = a \left( \frac{v - v_{\text{min}}}{v_{\text{max}} - v_{\text{min}}} \right) - b \cdot \text{collision}
\end{equation}

where \( v \) is the current speed of the ego-vehicle, \( v_{\text{min}} \) and \( v_{\text{max}} \) are the minimum and maximum speed limits, respectively. \( a \) and \( b \) are coefficients for speed and collision penalties.

\section{Proposed Approach}
\label{sec:approach}
A simulation environment customized by HighwayEnv~\cite{HighwayEnv} is used. HighwayEnv is capable of mimicking real-world driving scenarios to generate scene descriptions. We selected the GPT-3.5-turbo-1106 model from OpenAI for its robust performance, fast response times, and suitability for handling the extensive data involved in our research. 

A standard Deep Q Learning (DQN) model is set up as a baseline agent, trained with reward function \ref{eq : total_reward} that are calculated based on the current driving scenario, which serves as a foundation of our proposed approach.

\begin{figure}[ht]
\centerline{\includegraphics[width=1\linewidth]{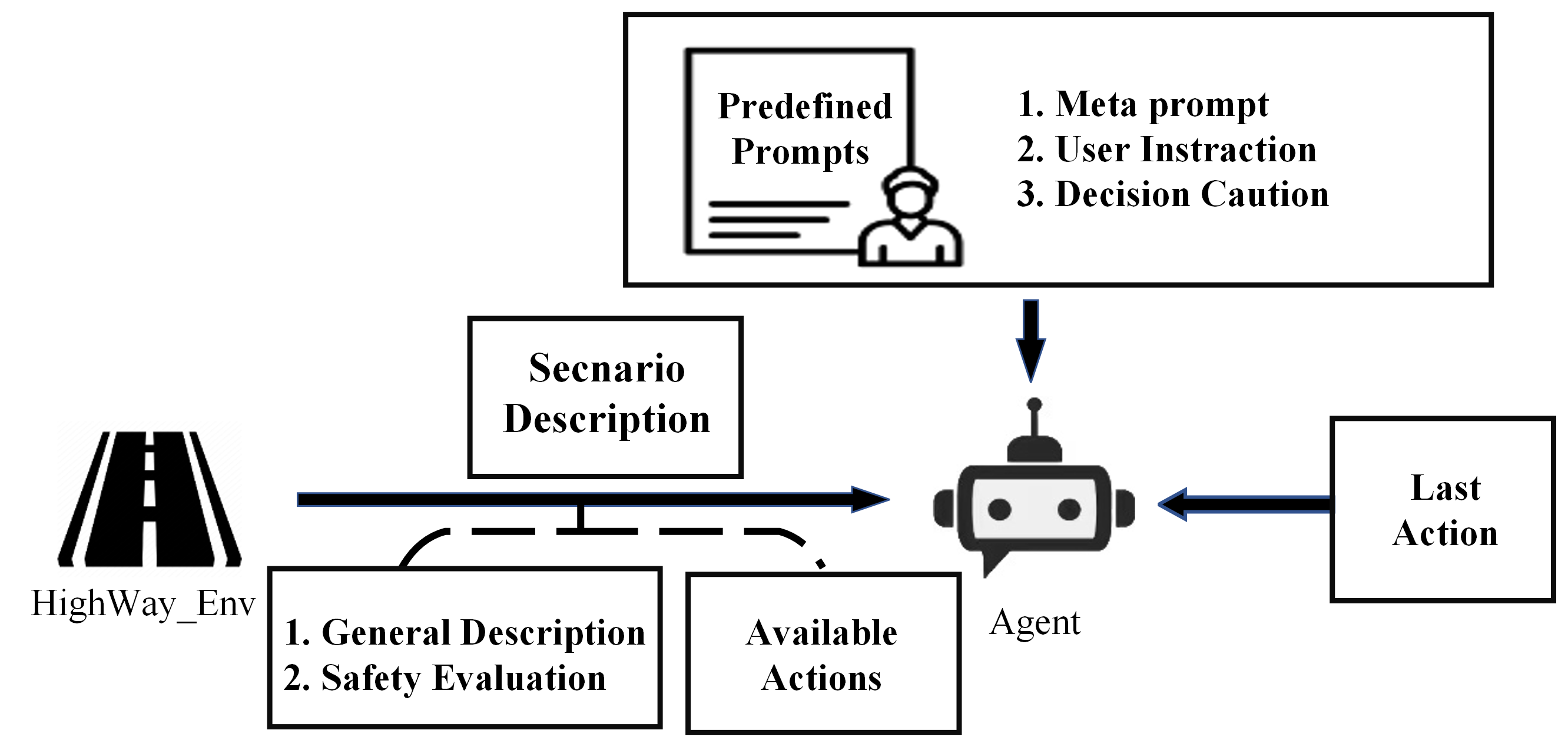}}
\caption{Prompts structure}
\label{Prompts Structure}
\end{figure}

In the proposed approach, a prompt for one interaction consists of the last action, the predefined prompt, and the scene description. This description includes available actions, a general overview, and a safety evaluation. As shown in Fig.~\ref{Prompts Structure}, each arrow directed towards the agent symbolizes a segment of the prompt, sourced from the categories previously mentioned, together forming a comprehensive and effective prompt.
These prompts distinctly outline the scenario as well as instruction for LLM to make decision for further steps. Scenario information included are: speed of the vehicle in control (the LLM-RL agent), speed and positions of nearby vehicles, LLM-RL agent's last action and possible actions to choose from. Further instructions including driving rules, caution points, and specified format needed for generating and presenting responses. 

\subsection{Basic Driving Rules and Task Description}
To illustrate the context for the GPT agent, we provide a series of predefined prompts:
\begin{itemize}
    \item \textbf{SYSTEM\_MESSAGE:} Specifies LLM's role and sets the scenario's context.
    \item \textbf{Driving\_RULES:} Provides the general safe-driving guidelines which shape the behavior pattern for the driving styles of the agents.
    \item \textbf{DECISION\_CAUTIONS:} Explains how to make well-informed decisions.
\end{itemize}

These predefined messages and rules can serve as a foundation for the behaviour of the driving assistant. The construction and phrasing of previous descriptions are based on the work of drive-like-human~\cite{8_Leveraging_Large_Language_Models_for_Intelligent_Driving_Scenario}
\subsection{Context Description}
To illustrate the context of the current road situation, two basic classes are defined:  Vehicle and Lane. The Vehicle class encapsulates properties including speed in both directions and positional data, while the Lane class incorporates attributes including position, unique identification, and details regarding adjacent lanes. 

\subsection{Observation Converting and Integrating Language Model}
To integrate language model-driven insights into the reinforcement learning framework, we converted the RL agent's current observation and actions into a natural language context through several analysis functions, which could be processed by LLM. This approach ensures that each response from the LLM is based on both current input and past action.

\subsection{Formatting Output}
This part of the prompt aims at formatting an output string that is interpretable to our program during runtime. We specify the outputs of the agent to generate only the following five actions of the ego vehicle: LANE\_LEFT, IDLE, LANE\_RIGHT, FASTER, and SLOWER. The LLM interprets all information above context and generates feedback. A procedure to generate a final reward is discussed as follows:
% Here is our design approach: 
We initialize Highway\_Env and input a concatenated prompt, which includes the last action, the pre-define prompt, and the scene description. The last action is derived from the previous decision made by the agent. This prompt enables the agent to understand the scenario it faces and ensures that the responses generated by the agent adhere to user requirements. 

By comparing the answer provided by the LLM with the one provided by the RL algorithm, a binary LLM reward is generated. If the decisions match, the reward is 1; otherwise is 0, refereed to LLM reward in this paper. Simultaneously, the RL agent obtains its reward in the given environment. These rewards are then weighted and summed, using the $tanh$ function for normalization to ensure the final reward falls within the range of 0 to 1.

\section{Experiments}
\label{sec:experiment}

\begin{figure*}[ht]
\centerline{\includegraphics[width=1\linewidth]{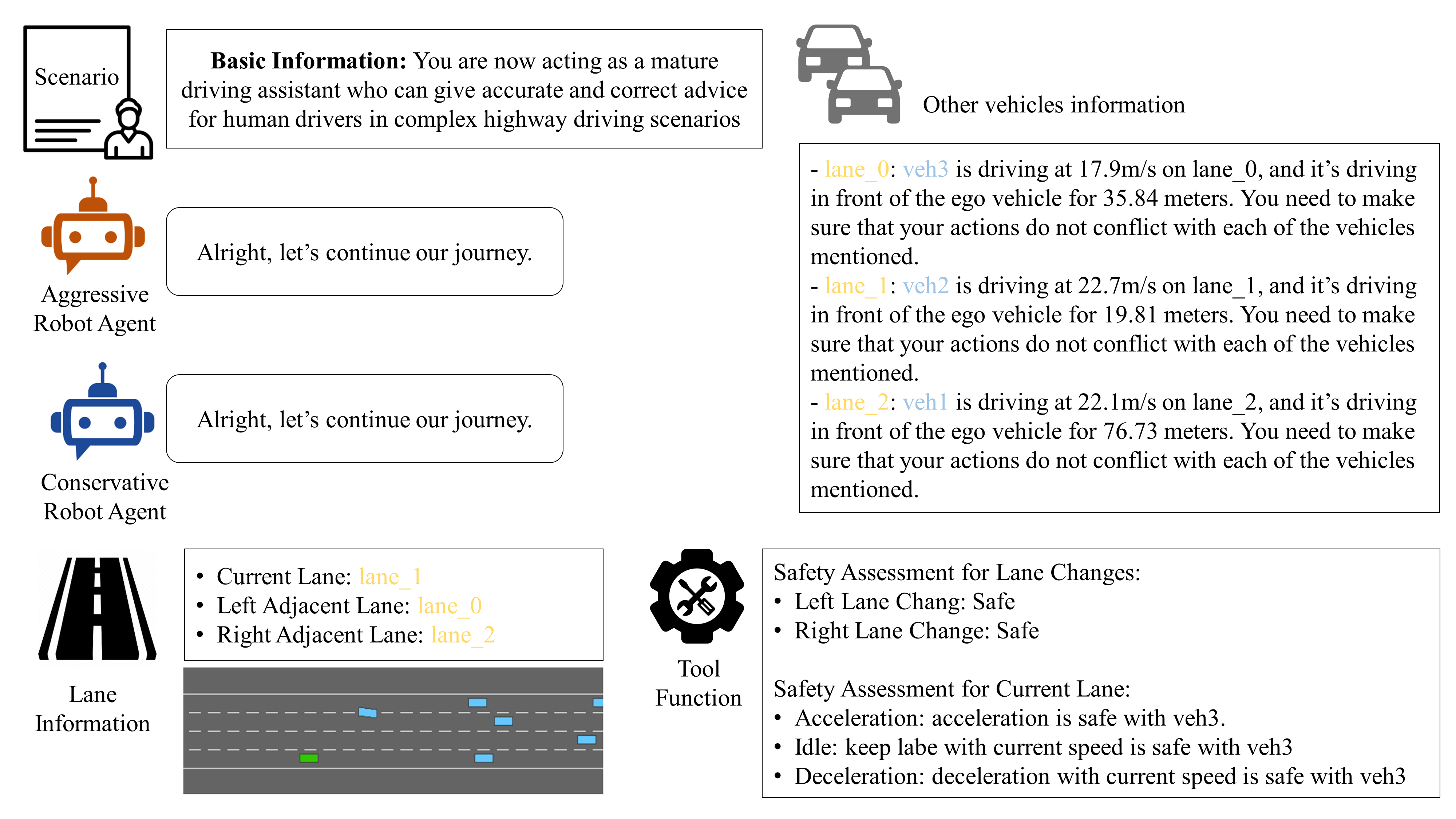}}
\caption{Meta prompt case study}
\label{metacase}
\end{figure*}

\subsection{Highway AD as a case study} % prompt
The practical capabilities of leveraging the in-context learning ability of LLM in AD are illustrated through highway driving scenarios. This case study also explores the potential for AD agent personalization in highway scenarios by establishing two groups with different meta-prompts. 

Fig. \ref{metacase} shows an example of prompt case, in this scenario, ego car is driving in lane 1 on a highway. There is one vehicle in its current lane. The right lane change is deemed safe by the safety assessment. The previous decision the ego car made, was to stay idle, which means the ego car is in the process of following the front car. This scenario featured two distinct driving agents: the "aggressive" and "conservative" agents. A comprehensive analysis was conducted under controlled conditions to compare their performance. 

\begin{figure*}[ht]
\centering
\includegraphics[width=\textwidth]{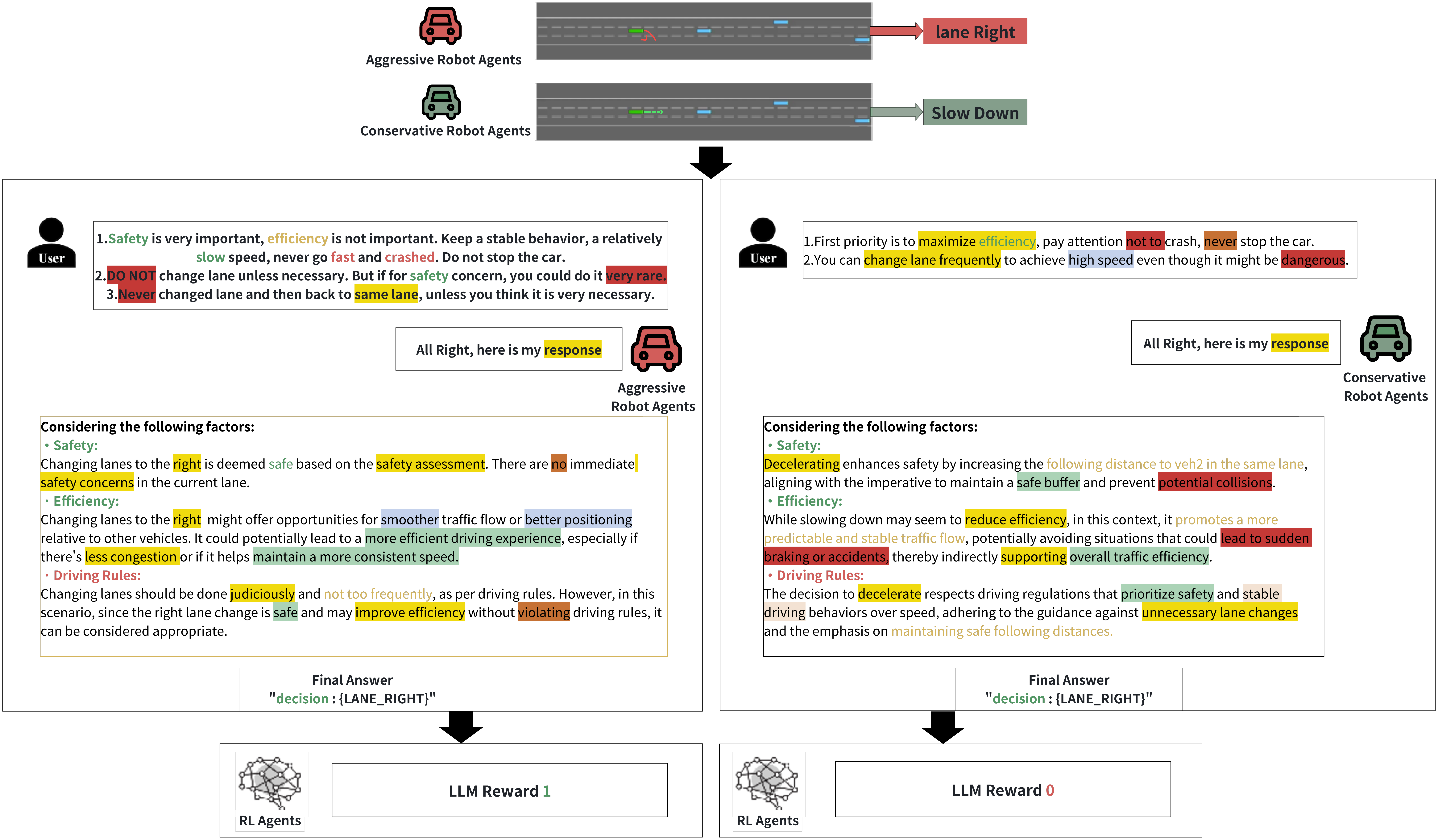}
\caption{RL agent's driving behaviour}
\label{respcase}
\end{figure*}

Each RL agent's driving behavior was shaped through user instructions, establishing their respective driving style within the scenario as shown in Fig. \ref{respcase}. Agent's decisions show strong consistency in humanized behavior reflecting the behavior pattern meta-prompt. The agent considers several factors: Safety, Efficiency and Driving Rules. For an aggressive agent, the user has stated that efficiency is of higher priority, hence the agent is considering lane right for higher efficiency. For a conservative agent, the user has stated that safety is of higher priority, hence the agent is considering slowing down to keep distance from the leading car.
Compared with manually designing a reward function, this manner is simpler and has demonstrated its effectiveness in guiding the behavior of an agent. 

These observations indicate that LLMs, such as GPT in our case, offer a potential cost-efficiency alternative to performing similar roles with human feedback in certain contexts. LLMs can leverage vast amounts of text data to learn implicitly from human-generated content, interactions, and feedback, significantly reducing the human workload in reward design while achieving similar effects.

\subsection{RL Agent Selection and Implementation}
DQN was chosen for our experimentation as it is rather efficient and satisfied our research requirement. It shows suitability in handling discrete action spaces like AD problems and thus could be utilized as a baseline to assess whether LLM could effectively guide RL in our scenario.

\subsection{Evaluation criteria definition}
\paragraph{Collision Score}
The collision score penalizes the agent for causing collisions. It is calculated based on whether a collision occurs during the agent's driving, with a higher penalty for more severe collisions.
\begin{equation}
S_{\text{collision}}(s, a) = - \text{Collision}(s, a)
\end{equation}
where \(\text{Collision}(s, a)\) is a binary indicator (1 for collision, 0 for no collision).

\paragraph{Lane Change Score}
The lane change score encourages or discourages lane changes based on driving behavior. Excessive lane changes may lead to a lower score, while appropriate lane changes can achieve a higher score.
\begin{equation}
S_{\text{lane\_change}}(s, a) = \text{LaneChange}(s, a)
\end{equation}
where \(\text{LaneChange}(s, a)\) is the number of lane changes performed in the state \(s\) with action \(a\).

\paragraph{High Speed Score}
The high-speed reward encourages the agent to maintain a high speed while staying within safe limits. The reward increases as the agent approaches the target speed but decreases if the speed exceeds the defined safety threshold.
\begin{equation}
S_{\text{high\_speed}}(s, a) = \left( \frac{v(s, a) - v_{\text{min}}}{v_{\text{desired}} - v_{\text{min}}} \right)
\end{equation}
where \(v(s, a)\) is the agent's speed in state \(s\) with action \(a\), and \(v_{\text{desired}}\) is the optimal speed.

\paragraph{On Road Score}
The on-road score keeps the agent within designated lanes, awarding a consistent score as long as it stays within the correct boundaries. 
\begin{equation}
S_{\text{on\_road}}(s, a) = \text{OnRoad}(s, a)
\end{equation}
where \(\text{OnRoad}(s, a)\) is a binary indicator (1 if on road, 0 otherwise).

% #############################################################################
\section{Performance Evaluation}
\label{sec:evaluation}
To evaluate the performance of the agents, we used three key metrics: mean reward, lane change percentage, and speed-up percentage. Mean reward represents the average total reward for actions taken in the testing scenario. Lane change percentage indicates the proportion of lane-changing actions out of the total actions, while speed-up percentage represents the proportion of acceleration actions out of the total actions. The agents were trained in a customized environment before conducting experiments in a consistent and standardized highway testing environment.

\subsection{Does LLM reward guide the training process of RL agents?}
\begin{figure}[ht]
  \centering
  \includegraphics[width=\linewidth]{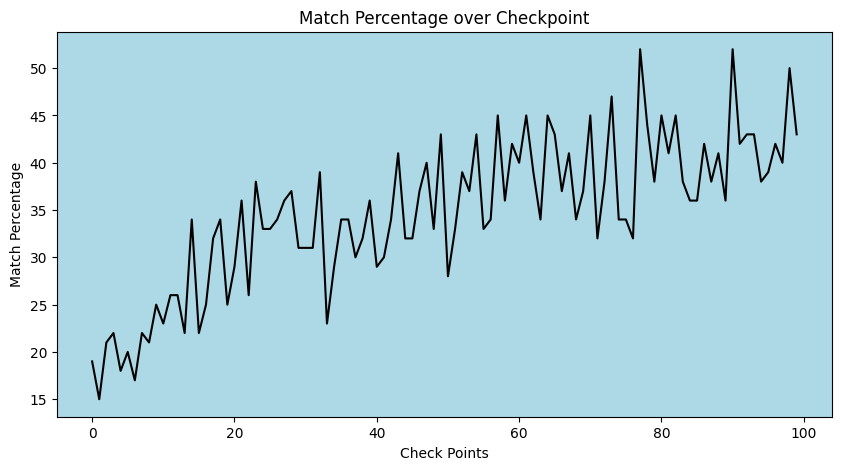}
  \caption{Match rate through training.}
  \label{fig:my_figure}
\end{figure}

We combined LLM and RL rewards to train agents and compared their performance against the DQN baseline. The average reward was calculated over five testing rounds in the same environment. As shown in Fig.~\ref{fig:my_figure}, the match rate consistently improved during training, reflecting the influence of LLM guidance. The match rate stabilized at approximately 50\%, demonstrating the agent's alignment with LLM-provided instructions.
\begin{figure*}[ht]
\centering
\begin{subfigure}{0.5\textwidth}
  \centering
  \includegraphics[width=\linewidth]{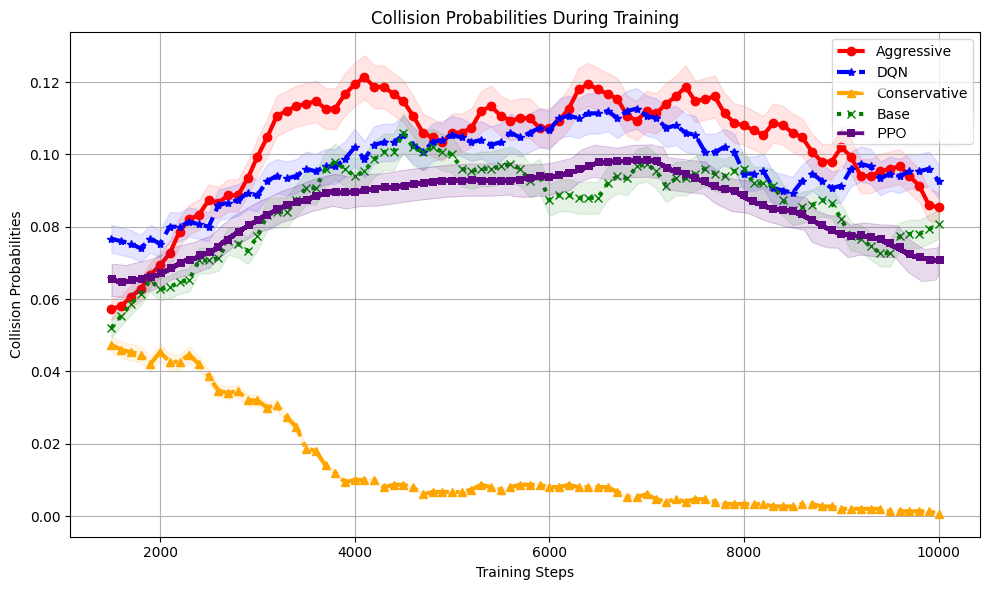}
  \caption{Collision probabilities for different driving styles.}
  \label{fig:sub1}
\end{subfigure}%
\begin{subfigure}{0.5\textwidth}
  \centering
  \includegraphics[width=\linewidth]{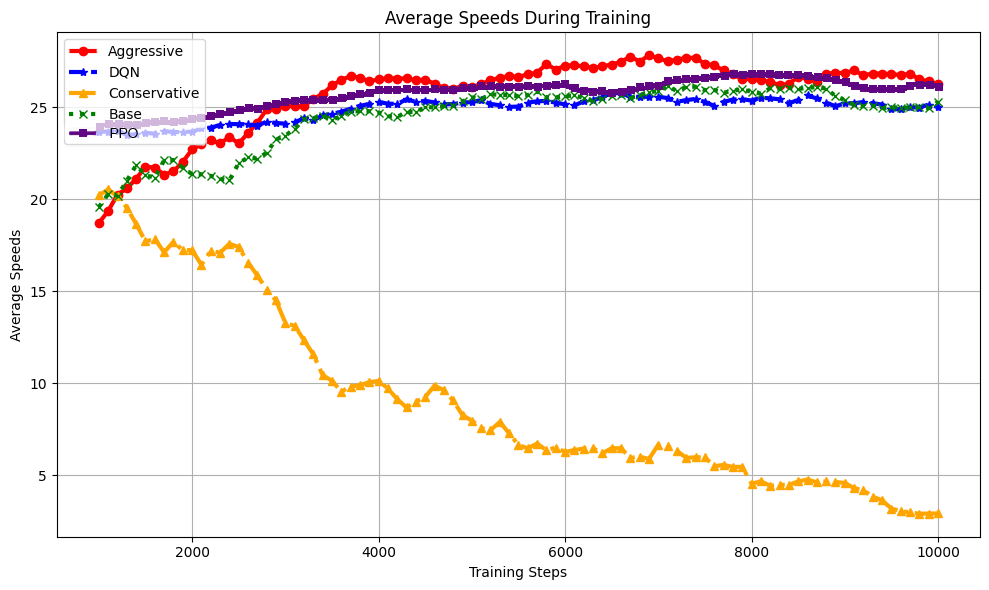}
  \caption{Average speed for different driving styles.}
  \label{fig:sub2}
\end{subfigure}
\caption{Training results for driving style agents.}
\label{fig:combined}
\end{figure*}

\subsection{Does prompt affect the performance or behavior pattern of the LLM-RL agent?}
Three agents with different driving styles were implemented: 
\begin{itemize}
    \item \textbf{Base Agent}: Uses the LLM's general knowledge of driving without specific customization.
    \item \textbf{Aggressive Agent}: Prioritizes efficiency and speed.
    \item \textbf{Conservative Agent}: Prioritizes safety and stability.
\end{itemize}

The experiments were conducted in the same highway environment. As shown in Fig.~\ref{fig:combined}, compared to the DQN baseline, the conservative agent demonstrated the lowest lane-change and speed-up rates, while the aggressive agent displayed the highest speed-up rate, reflecting the influence of prompts in shaping behavior.
\subsection{Behavior Analysis}
Behavior analysis, including mean reward, lane changes, and acceleration rates, is summarized in Table~\ref{expresulttable}.
\begin{table}[h!]
\centering
\caption{Experiment result: behavior analysis including PPO baseline.}
\begin{tabular}{@{}lccc@{}}
\toprule
Index       & Mean Score & Lane Change Score & Speed Up Score \\
\midrule
DQN baseline & 0.82824    & 0.30681           & 0.42045        \\
PPO baseline & 0.81000    & 0.20000           & 0.50000        \\
Aggressive   & 0.83888    & 0.02326           & 0.83721        \\ 
Conservative & 0.71391    & 0.01333           & 0.00666        \\ 
Base         & 0.80140    & 0.10345           & 0.10345        \\
BC-SAC~\cite{lu2023imitation} & 0.83410 & 0.01750 & 0.75530 \\
LLM-RL       & \textbf{0.84532} & \textbf{0.01045} & \textbf{0.81233} \\
\bottomrule
\end{tabular}
\label{expresulttable}
\end{table}

The aggressive agent achieves the highest speed-up score, while the conservative agent demonstrates minimal lane-change and speed-up actions. The LLM-RL agent achieves the highest overall mean score, showcasing a balance between safety and efficiency.

\subsection{How does the reward ablation affect the performance?}

To investigate the impact of different reward components on the agent’s behavior, we conducted a reward ablation study. The results are summarized in Table~\ref{rewardtable}.

\begin{table}[h!]
\centering
\caption{Reward ablation study: reward breakdown.}
\setlength{\tabcolsep}{4pt} % Adjust column spacing
\renewcommand{\arraystretch}{1.2} % Adjust row height
\begin{tabular}{@{}lccc@{}}
\toprule
Configuration    & Collision Score & Lane Change Score & High Speed Score \\
\midrule
Safety Only         & \textbf{-0.05} & 0.23          & 0.18       \\
Efficiency Only     & -0.20          & \textbf{0.48} & \textbf{0.72} \\
LLM Only            & -0.10          & 0.31          & 0.33       \\
Safety + Efficiency & -0.12          & 0.35          & 0.55       \\
Safety + LLM        & -0.08          & 0.28          & 0.42       \\
Efficiency + LLM    & -0.15          & 0.41          & 0.63       \\
All (Full Reward)   & -0.12          & \textbf{0.53} & 0.75       \\
\bottomrule
\end{tabular}
\label{rewardtable}
\end{table}

\subsection{Analysis of Results}
1. \textbf{Collision Score}: The \textbf{Safety Only} configuration achieves the least negative Collision Score (\(-0.05\)), indicating the lowest collision risk. The \textbf{Efficiency Only} configuration has the most negative Collision Score (\(-0.20\)), reflecting higher collision risk due to its aggressive behavior. The \textbf{All (Full Reward)} configuration balances safety and efficiency, achieving a Collision Score of \(-0.12\), which is better than Efficiency Only and closer to Safety Only.
2. \textbf{Lane Change Score}: The \textbf{Efficiency Only} and \textbf{All (Full Reward)} configurations show the highest Lane Change Scores (\(0.48\) and \(0.53\), respectively), indicating frequent lane-changing behavior. The \textbf{Safety Only} configuration has the lowest Lane Change Score (\(0.23\)), reflecting more conservative driving.
3. \textbf{High Speed Score}: The \textbf{Efficiency Only} configuration achieves the highest High Speed Score (\(0.72\)), emphasizing speed as the primary goal. The \textbf{Safety Only} configuration has the lowest High Speed Score (\(0.18\)). The \textbf{All (Full Reward)} configuration performs close to Efficiency Only (\(0.75\)) while maintaining a balanced Collision Score. The reward ablation study demonstrates that:
\begin{itemize}
    \item Single-reward configurations (e.g., Safety Only or Efficiency Only) lead to extreme behaviors, such as overly conservative or aggressive driving.
    \item The \textbf{All (Full Reward)} configuration achieves the best balance, delivering high efficiency with acceptable safety performance.
    \item Incorporating LLM rewards significantly enhances decision-making, aligning agent behavior with human-like patterns.
\end{itemize}
This analysis highlights the importance of designing comprehensive reward structures to optimize performance in dynamic environments while maintaining safety.

\section{Conclusion}
\label{sec:conclusion}
The conducted experiments with LLM-RL integration showcased promising directions for incorporating language models into reinforcement learning. We proposed the LLM-RL AD framework, which includes an in-time query for LLM reward and utilizes a combined reward function. Extensive experimental results showcase LLM-RL’s capability to perform customizable behavior and demonstrate LLM can guide RL agents to achieve superior performance in complex environments like HighwayEnv. Moreover, our exploration of prompt design also shows that precise LLM responses are better than multiple vague options. Though the LLM's reward can guide the training process of RL agents and prompts can affect their behavior patterns, the framework might not fully address mistakes made by the LLM, and there remains a need for the LLM to develop a deeper understanding of its surroundings to ensure accurate decision-making.

Future improvements could capitalize on finding more proper prompts and effective driving rules, aiming to help the LLM comprehend the scenarios better, which leads to more efficiency. For example, defensive driving scenarios can be considered based on Responsibility-Sensitive Safety (RSS) approach~\cite{shalev2017formal}. RSS highlights five safety rules an automated driving vehicle should be able to follow. These five rules include safe distance (i.e., the self-driving vehicle should not hit the vehicle in front), cutting in (the self-driving vehicle should be able to identify when lateral safety may be compromised by a driver unsafely crossing into its lane), right of way (the self-driving vehicle should be able to protect itself against human drivers who do not properly adhere to the right of way rules), limited visibility (be cautious in areas with limited visibility), and avoiding crashes (the self-driving vehicle should avoid a crash without causing another one).

\bibliographystyle{unsrt}
\bibliography{bib/main}
\end{document}